\documentclass[10pt,twocolumn,letterpaper]{article}

\usepackage[pagenumbers]{wacv} 

\usepackage{pifont}
\definecolor{wacvblue}{rgb}{0.21,0.49,0.74}
\usepackage[pagebackref,breaklinks,colorlinks,allcolors=wacvblue]{hyperref}

\def\wacvPaperID{***} 
\def\confName{WACV}
\def\confYear{2027}

\title{GraspIT: A Dataset Bridging the Sim-to-Real gap and back for Validated Grasping SE(3) Pose Generation}

\author{Paul Koch\\
Fraunhofer IPK\\
Pascalstraße 8-9, 10587 Berlin, Germany\\
{\tt\small paul.koch@ipk.fraunhofer.de}
\and
Adem Karakurt\\
Fraunhofer IPK\\
\and 
André Sers\\
Fraunhofer IPK\\
}

\begin{document}
\maketitle
\begin{abstract}
Robust robotic grasping of novel objects requires datasets that simultaneously provide photorealistic RGB-D observations, physically validated grasp quality annotations, and a principled bridge between simulation and the real world, which existing datasets lack to provide jointly. \textbf{GraspIT} addresses this gap: tabletop scenes in NVIDIA  Isaac Sim are annotated via a four-stage physical slip-test on parallel Franka Panda instances, producing trajectory-reachability checks and continuous quality scores beyond force-closure.Of ${\sim}$2.3M candidates, 83\% pass as \emph{good} ($s{\geq}0.50$); the 17\% that passed force-closure but failed the slip-test provide graded hard negatives. A Real$\leftrightarrow$Sim loop back-projects these labels onto 100 real-world scenes. The release provides ${\sim}$316k annotated RGBD frame sets across 1\,035 sim and 100 real scenes, with instance masks, 6-DoF poses, physical object properties, and scored 6-DoF grasps. All tools are open-source and Docker-containerized. The trajectory planning within Isaac Sim further allows streaming of high resolution demonstrations for tabletop manipulation policy learning and behavior cloning. 
\end{abstract}

\section{Introduction}
\label{sec:intro}
Robotic manipulation of everyday objects --- picking, placing, stacking, and handing over --- remains an unsolved challenge despite decades of research~\cite{FasterRCNN,GraspNetDs}. Recent large-scale imitation learning systems such as RT-2~\cite{rt2}, $\pi_0$~\cite{pi0_paper}, and Octo~\cite{octo} demonstrate multi-task instruction-following in laboratory settings; yet, their generalization collapses when confronted with objects, scenes, or lighting conditions outside the training distribution. A central root cause is the absence of \emph{physically grounded} visual representations: the encoders driving these systems are trained on semantic similarity objectives (CLIP~\cite{clip}, SigLIP~\cite{siglip2}) that do not encode metric geometry, 3D object understanding, or the physical grounding that determines whether a proposed grasp or object manipulation will succeed~\cite{knowledgeinsulation, yang2026GeoVLAs}.

\begin{figure}
    \centering
    \includegraphics[width=0.5\linewidth]{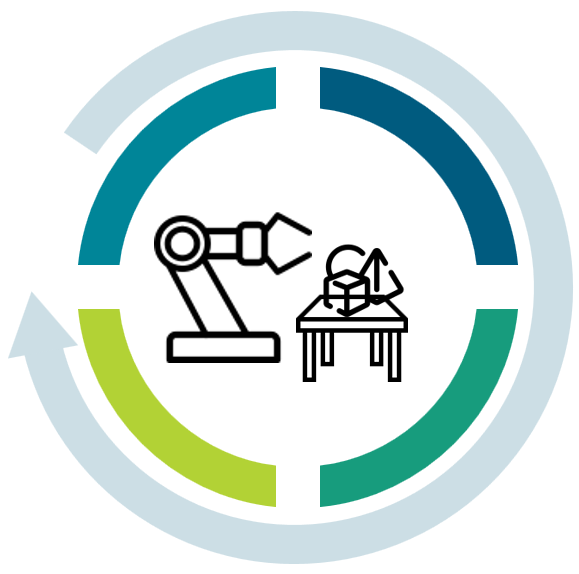}
    \caption{\textbf{Grasp} \textbf{IT}erative robot tabletop object manipulation dataset: illustration of the four-stage iterative concept.}
    \label{fig:logo}
\end{figure}
Collecting the training data needed to close this gap is expensive. Accurate 6-DoF grasp annotation requires either real-robot execution at scale or a physics simulator; real RGBD observations necessitate a physical setup; linking the two so that simulation annotations are valid for real sensor data requires a careful calibration pipeline. No existing dataset provides all of these simultaneously (see Tab.~\ref{tab:dataset_landscape}).

We introduce \textbf{GraspIT} (Fig.~\ref{fig:logo}), a dataset and generation system that addresses this gap through four principled design decisions (Fig.~\ref{fig:real_branch_capture_pipeline}): \textbf{1) Quality over quantity.} Every grasp candidate is validated by a four-stage physical \emph{slip test} executed by a simulated Franka Panda arm (Fig.~\ref{fig:robotershool}), producing a continuous quality score $s \in \{0, 0.25, 0.50, 0.75, 1.0\}$ that captures failure modes (slip under gravity, oscillation, pendulum swing) that are absent from force-closure metrics. \textbf{2) Robot-in-the-loop reachability filtering.} Every candidate is validated for collision-free trajectory reachability by the robot arm \emph{before} the slip test, ensuring that all positive labels are actionable. \textbf{3) Real$\leftrightarrow$Sim bidirectional link.} Real-world scene geometry is captured via a robot-mounted camera and registered into Isaac Sim; simulation-validated labels are back-projected onto all real camera frames. This means a single real-world capture session yields both real RGB-D images and unlimited additional photorealistic synthetic views with full annotation (Fig.~\ref{fig:real_branch_scan}). \textbf{4) Open, containerized, and extendable.} All generation tools are packaged as Docker containers, enabling any lab to reproduce, extend, or contribute to the shared GraspIT data foundation.

GraspIT initial release provides $\sim316$k annotated RGBD frame sets across 1\,035 Isaac Sim scenes and 100 real-world scenes, with ${\sim}$2.3M slip-test validated scene unique grasp candidates spanning ${\sim}7.4$k+ unique object identities drawn from the ABC dataset~\cite{abcDS} and our own real assets. Every frame carries per-instance segmentation masks, 6-DoF object poses, scored visible/hidden grasping poses, camera calibrations, complete 3D surface point clouds from CAD, analytically derived physical properties (centroid, extent, volume), and robot trajectories to grasping attempts. 

\paragraph{Contributions.} \textbf{(A)} GraspIT: the first dataset providing photorealistic RGBD observations, robot-reachability-filtered grasps, and staged slip-test quality scores simultaneously. \textbf{(B)} A Real$\leftrightarrow$Sim registration pipeline that makes simulation annotations valid for real sensor data. \textbf{(C)} A scalable, open, containerized generation system enabling unlimited new scene generation with minimal manual labeling. 

\section{Related Work}
\label{sec:relatedwork}
\paragraph{6-DoF Object Pose Datasets (generatable object properties).} The BOP benchmark~\cite{BoP23} consolidates eight pose estimation datasets (LineMOD~\cite{LineMod}, YCB-Video~\cite{PoseCNN}, T-Less~\cite{T-Less}, HomebrewedDB~\cite{HomeBrew}, and others), providing calibrated RGBD frames with CAD-model annotations and 6-DoF ground truth. These datasets are invaluable for detection and shape reasoning but carry \emph{no manipulation annotations}: grasping labels, robot reachability, and physical object properties are absent. The MVIP dataset~\cite{MVIP} extends benchmarking to industrial multi-view settings with 308 object classes, including real object weight and language property annotations —-- but similarly focuses on recognition rather than manipulation.

\paragraph{Grasping Datasets.} Cornell~\cite{cornellgrasp} and Jacquard~\cite{jacquard} provide 2D planar grasp rectangles --- a useful simplification but incapable of representing tilted or sideways approach configurations required in real manipulation. 

GraspNet-1Billion~\cite{GraspNetDs} is the dominant 6-DoF benchmark, providing real RGBD tabletop imagery with 1B analytically scored grasp candidates. However, its \emph{force-closure metric}~\cite{GraspNetDs} assumes rigid, perfectly known contact geometry, which has been shown to correlate poorly with real-robot success under dynamic perturbation~\cite{mahler2017dexnet}. Graspnet further reports only a moderate grasping success correlation between a subset of ``good'' grasp candidates and tested robot success. Critically, GraspNet carries no robot-reachability filtering: many high-scoring candidates require the arm to reach through the support surface or exceed workspace limits. 

ACRONYM~\cite{acronym} advances beyond analytic scoring by executing 17.7M grasp candidates in NVIDIA FleX under gravity-free shaking, but it provides \emph{no visual observations}, uses non-textured ShapeNet meshes~\cite{shapenet} that produce a large sim-to-real gap, and does not model robot kinematics. GraspIT directly addresses all three limitations.

\paragraph{Learning-Based Grasp Quality Estimation.} Dex-Net~2.0~\cite{mahler2017dexnet} introduced the Grasp Quality Convolutional Neural Network (GQ-CNN), trained on synthetic depth images paired with analytic grasp-quality metrics to predict parallel-jaw grasp success from a single depth image. While demonstrating strong sim-to-real transfer for planar grasps, Dex-Net~2.0 relies on force-closure as its simulation ground truth, inheriting the correlation gap between analytic scores and real execution that motivates our slip-test design. Dex-Net~4.0~\cite{mahler2019dexnet4} extended this to ambidextrous policies combining suction and parallel-jaw grasping, reporting ${\sim}9$--$98\%$ real-robot success depending on quality score --- the correlation range that motivates GraspIT's staged validation. Contact-GraspNet~\cite{sundermeyer2021contactgraspnet} moves beyond quality prediction to direct 6-DoF grasp \emph{generation} from point clouds in cluttered scenes, using GraspNet-1B as its training corpus. Its strong real-world performance highlights the value of scene-level RGBD observations --- a modality absent from ACRONYM and present in GraspIT. Critically, none of these methods validates candidate reachability via robot trajectory planning, nor provides a Real$\leftrightarrow$Sim registration link; GraspIT is designed to supply exactly these missing components as a pretraining foundation for next-generation object manipulation models.

\paragraph{Robot Demonstration Datasets.} Large-scale imitation learning datasets --- BridgeData V2~\cite{bridgedatav2}, DROID~\cite{droid}, and Open X-Embodiment~\cite{openx} --- provide paired observation-action sequences and have driven the development of generalized manipulation policies~\cite{rt1,rt2,pi0_paper}. However, they carry no explicit 3-D geometry, CAD models, or grasp quality scores; they teach \emph{what to do} but not \emph{why it works physically}. ManiSkill2~\cite{maniskill} provides simulation-based demonstrations with object metadata but focuses on task-success trajectories rather than physical object properties or the Real$\leftrightarrow$Sim link. GraspIT is positioned as a complementary pretraining foundation whose physically grounded labels can improve the visual representations consumed by VLAs and diffusion policies~\cite{diffusionpolicy, dp3, knowledgeinsulation}.

\paragraph{Geometry-Aware Visual Pretraining.} Recent work has shown that explicitly injecting geometric knowledge into visual encoders improves downstream robotic perception. Here, Yang \etal~\cite{yang2026GeoVLAs} recently quantified a $2{\times}$ depth-estimation gap between the VGGT geometric foundation model and the GR00T-N1.5 VLA's encoder, confirming that semantic pretraining leaves a measurable physical reasoning deficit. GraspIT is the first dataset designed to support pretraining that fills this object-level physical grounding gap with directly manipulable, physics-validated labels for robot tabletop manipulation tasks.

\paragraph{Sim-to-Real Data Generation.} Synthetic data generation with domain randomization~\cite{domainrand} and photorealistic rendering~\cite{blenderproc, rasim} has become standard practice for pose estimation and detection training. Robot-assisted annotation pipelines~\cite{KOCHRobotAssited6D} demonstrate that real-world 6-DoF labels can be collected with minimal user interaction. Our Real$\leftrightarrow$Sim loop extends this philosophy to grasp quality: annotations generated in simulation are explicitly validated for transfer to real sensor frames through metric 3D registration, addressing the domain gap documented in the BOP challenge~\cite{BoP23}.

\begin{table*}[t]
  \centering
  \caption{\textbf{Dataset Landscape with Robot Manipulation relevant data modalities and training targets.} 
  Columns: 
  \textbf{A}~Metric depth/scene geometry; 
  \textbf{B}~Per-instance segmentation; 
  \textbf{C}~6-DoF pose + CAD; 
  \textbf{D}~Generated Object Properties (shape extend, center-of-mass, visibility, volume, scale, position); 
  \textbf{E}~Physically validated grasps; 
  \textbf{F}~Hierarchical slip scores; 
  \textbf{G}~Photorealistic assets; 
  \textbf{H}~Real$\leftrightarrow$Sim link; 
  \textbf{I}~Robot demonstrations. 
  }
  \label{tab:dataset_landscape}
  \begin{tabular}{l|c|r|r|ccccccccc}
    \toprule
    \textbf{Dataset} & \textbf{Year}
      & \textbf{Scenes} & \textbf{Frames}
      & \textbf{A} & \textbf{B} & \textbf{C}& \textbf{D}
      & \textbf{E} & \textbf{F} & \textbf{G} & \textbf{H} & \textbf{I} \\
    \midrule
    NYU-Depth V2~\cite{nyuDepthV2}     & 2012 & 464    & 1\,499   & \ding{51} & \ding{51} & \ding{55} & \ding{55} & \ding{55} & \ding{55} & \ding{55} & \ding{55} & \ding{55} \\
    LineMod~\cite{LineMod}             & 2013 & 15     & 15k      & \ding{51} & \ding{51} & \ding{51} & $\approx$ & \ding{55} & \ding{55} & \ding{55} & \ding{55} & \ding{55} \\
    YCB-Video~\cite{PoseCNN}           & 2018 & 92     & 134k     & \ding{51} & \ding{51} & \ding{51} & $\approx$ & \ding{55} & \ding{55} & \ding{55} & \ding{55} & \ding{55} \\
    GraspNet-1B~\cite{GraspNetDs}      & 2020 & 190    & 97k      & \ding{51} & \ding{51} & \ding{51} & $\approx$ & $\approx$ & \ding{55} & \ding{55} & \ding{55} & \ding{55} \\
    ACRONYM~\cite{acronym}             & 2020 & --     & --       & \ding{55} & \ding{55} & \ding{51} & $\approx$ & \ding{51}  & $\approx$ & \ding{55} & \ding{55} & \ding{55} \\
    ManiSkill2~\cite{maniskill}        & 2023 & 20+    & --       & \ding{51} & $\approx$ & $\approx$ & \ding{55} & \ding{55} & \ding{55} & \ding{55} & \ding{55} & \ding{51} \\
    Open X-Emb.~\cite{openx}          & 2023 & --     & 1M+ ep & \ding{51} & \ding{55} & \ding{55} & \ding{55} & \ding{55} & \ding{55} & \ding{55} & \ding{55} & \ding{51} \\
    DROID~\cite{droid}                 & 2024 & --     & 76k ep  & \ding{51} & \ding{55} & \ding{55} & \ding{55} & \ding{55} & \ding{55} & \ding{55} & \ding{55} & \ding{51} \\
    \midrule
    \textbf{GraspIT (ours)}            & 2026 & 1\,035 Sim + 100 Real & $\sim316$k & \ding{51} & \ding{51} & \ding{51} & \ding{51} & \ding{51} & \ding{51} & \ding{51} & \ding{51} & $\approx$ \\
    \bottomrule
    \multicolumn{13}{c}{\ding{51}~=~fully provided; $\approx$~=~partially or generate-able; \ding{55}~=~absent.}
  \end{tabular}
\end{table*}

\section{GraspIT: A Data Foundation for Object Manipulation}
\label{sec:graspit}

GraspIT is designed not as a static benchmark but as a \emph{living, expandable generation system}: a closed Real$\leftrightarrow$Sim loop (Fig.~\ref{fig:real_branch_capture_pipeline}) that produces arbitrarily large, fully annotated RGBD corpora from a single consistent pipeline. Its design is guided by the observation that downstream robot manipulation models suffer most from two compounding deficits—the lack of physically grounded pretraining data and the lack of a principled bridge between simulation labels and real sensor 
observations~\cite{yang2026GeoVLAs, knowledgeinsulation}. All tools are packaged as Docker containers and hosted at \url{https://github.com/KochPJ/GraspIt}.

\subsection{Design Philosophy}
\label{sec:philosophy}

GraspIT rests on four design pillars:

\paragraph{Quality over quantity.} Rather than generating billions of analytically scored candidates, GraspIT validates each through a four-stage physical slip test (Sec.~\ref{sec:robotschool}), yielding a continuous quality score $s \in [0,1]$. The base score $b$ is determined by the last stage successfully completed:

\begin{equation}
  b =
  \begin{cases}
    0.00 & \text{trajectory planning to closure failed} \\
    0.25 & \text{trajectory OK; lift failed} \\
    0.50 & \text{lift OK; horizontal oscillation failed} \\
    0.75 & \text{horizontal OK; pendulum swing failed} \\
    1.00 & \text{all four stages passed}
  \end{cases}
  \label{eq:base_score}
\end{equation}

The final quality score additionally penalizes excessive slip, even within a passed stage. For each stage $i$ that is executed, if the measured positional slip $|\Delta\mathbf{p}_i| > 2\,\text{cm}$ \emph{or} the measured orientational slip $|\Delta\boldsymbol{\theta}_i| > 10°$, a penalty of $\tfrac{1}{12}$ is subtracted:
\begin{equation}
  s = b \;-\; \frac{N_{\text{xyz}>2\text{cm}}}{12}
            \;-\; \frac{N_{\text{rpy}>10°}}{12},
  \qquad s \in [0, 1],
  \label{eq:quality_score}
\end{equation}
\noindent where $N_{\text{xyz}>2\text{cm}}$ (resp.\ $N_{\text{rpy}>10°}$) counts the number of executed stages in which the positional or orientational slip exceeds the threshold.  A grasp is annotated \textbf{good} if $s \geq 0.50$ and \textbf{bad} otherwise.

This two-level design is intentional: a grasp that nominally passes the lift stage but displaces the object by 3\,cm can be penalized below the ``good'' threshold, reflecting real-world unreliability.  The denominator~12 is calibrated so that a single maximal-slip violation on a 0.75-base grasp still keeps it marginally good ($ 0.75 - 1/12 \approx 0.67 $), while two violations on a 0.50-base grasp push it to bad ($0.50 - 2/12 \approx 0.33$). We clamp scores between $[0, 1]$ when reaching a theoretically reachable negative score. 

This design is motivated by the finding that Dex-Net~4.0's~\cite{mahler2019dexnet4} force-closure scores of $s{=}0.1$ yield only ${\sim}9$--$23\%$ real-robot success, whereas $s{=}1.0$ yields ${\sim}93$--$98\%$. Each stage of our slip test exposes a distinct failure mode absent from static force-closure analysis. 

\begin{figure*}[h]
  \centering
  \includegraphics[width=0.8\linewidth]{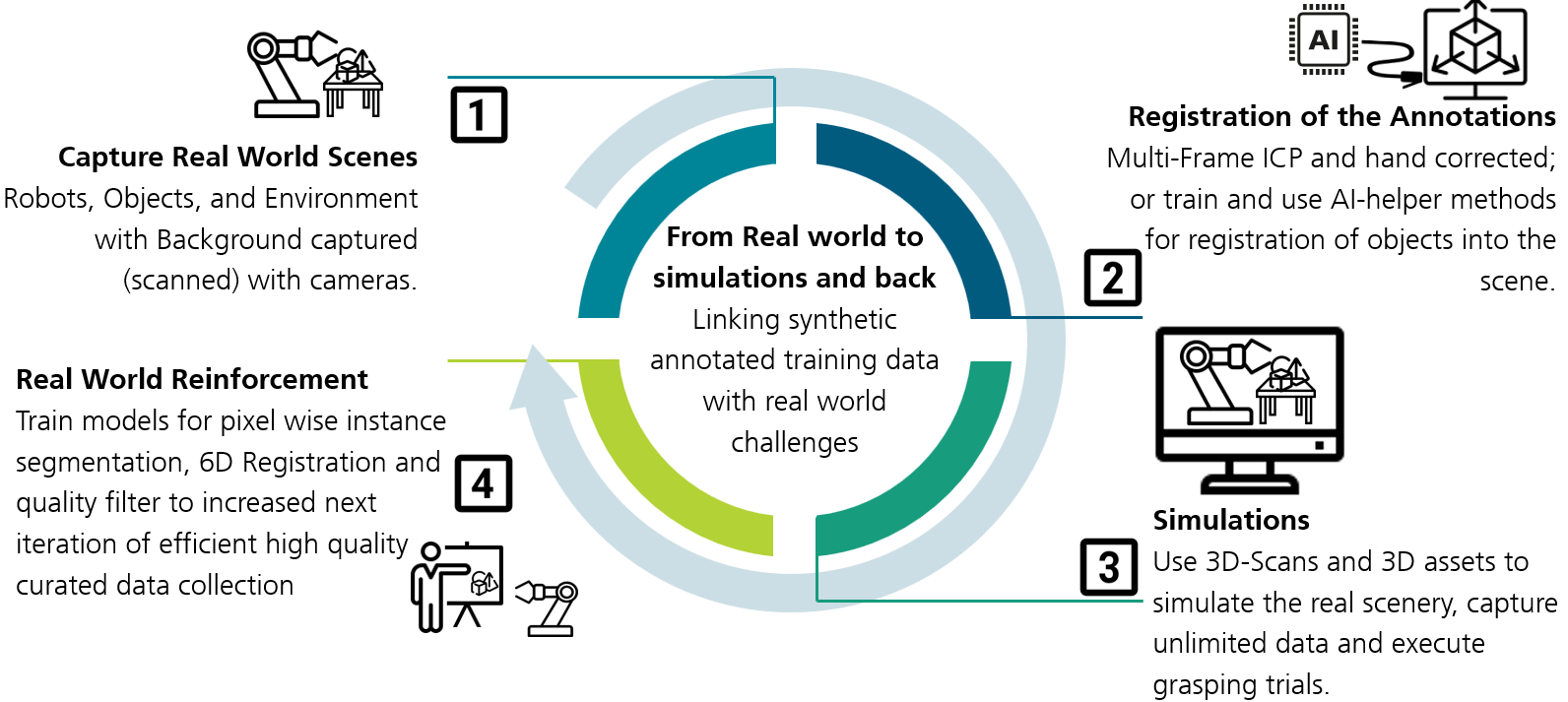}
  \caption{\textbf{GraspIT Real$\leftrightarrow$Sim Collection Loop.} Objects are 3-D scanned via generative image-to-3D reconstruction and imported into Isaac Sim. Real scenes are captured by a robot-mounted camera, registered into the simulator, and extended with synthetic views. Physics-validated grasp labels from the Robot School are back-projected onto all real and synthetic frames, yielding a consistent annotation coordinate frame across both domains.}
  \label{fig:real_branch_capture_pipeline}
\end{figure*}
\paragraph{Robot-in-the-loop reachability.} Every candidate passes collision-free trajectory planning by the Franka Panda arm via cuRobo~\cite{sundaralingam2023curobo} \emph{before} the slip test. Candidates requiring approaches through obstacles or beyond workspace limits are discarded, ensuring all positive labels are actionable for a deployed system.

\paragraph{Real$\leftrightarrow$Sim bidirectional link.} Real scene geometry is registered into Isaac Sim; simulation-validated annotations are back-projected to all real camera poses (see Fig.~\ref{fig:real_branch_capture_pipeline}). A single capture session therefore yields both real RGB-D images \emph{and} unlimited additional photorealistic synthetic views at novel viewpoints, all sharing a consistent annotation coordinate frame (Fig.~\ref{fig:real_branch_scan}).

\paragraph{Open and extendable infrastructure.} The Docker containerized pipeline accepts any registered scene, enabling the progressive integration of existing datasets (BOP, GraspNet) into the GraspIT framework and extension to new robot models, environments, or sensor modalities.

\subsection{Simulation Branch: Scene Generation}
\label{sec:simbranch}

\begin{figure*}[t]
  \centering
\includegraphics[width=\linewidth]{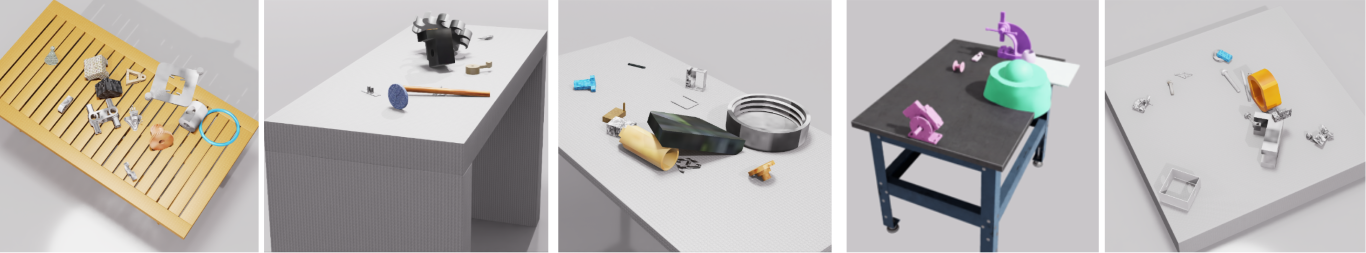} 
  \caption{\textbf{GraspIT Scene Variety.} Representative Isaac Sim tabletop scenes with ABC objects, randomised table textures, and HDR lighting. Every scene introduces novel object combinations.}
  \label{fig:scene_sampler}
\end{figure*}

Tabletop scenes are assembled in NVIDIA Isaac Sim~\cite{isaacgym} by a probabilistic scene sampler that draws 10--20 objects uniformly at random from the ABC dataset~\cite{abcDS} (${\sim}10$k accessible curated CAD meshes spanning industrial, organic, and household geometries). Each object per scene is randomly scaled to a maximum side length of $[2, 20]$\% of the maximum table side length—ensuring object scale diversity, graspable and non-graspable object instances, and disconnecting grasping candidates from object IDs. The sampled and scaled objects are then randomly oriented, spawned $[20, 40]$cm above the table, and dropped with gravity physics enabled to generate fully randomized, physically plausible scene configurations. We use ten table models with domain-randomized PBR textures, HDR lighting, and ambient occlusion, so each generated Sim scene presents a unique visual distribution (Fig.~\ref{fig:scene_sampler}).

For each scene, 256 camera poses are sampled from a hemispherical trajectory at 1--2\,m from the table center. Each frame is rendered at $1920{\times}1080$ and exported with: photorealistic RGB; metric depth with structured-light noise; per-instance semantic segmentation from the scene graph; ground-truth 6-DoF object pose; and complete CAD point cloud. Because of the large 10k ABC object pool, every new generation run introduces novel object combinations, preventing object-identity overfitting.

\subsection{Annotation Modalities}
\label{sec:modalities}
\begin{figure*}[h]
  \centering
  \begin{subfigure}{0.24\linewidth}
    \includegraphics[width=\linewidth]{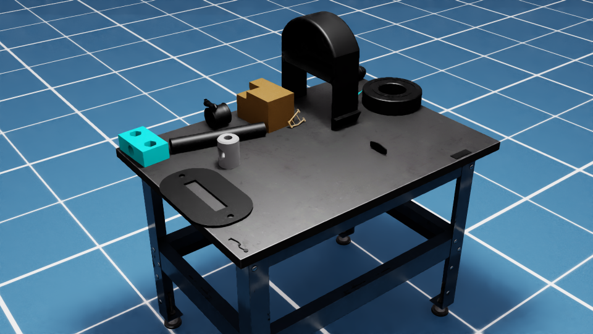}
    \caption{RGB}
  \end{subfigure}
  \begin{subfigure}{0.24\linewidth}
    \includegraphics[width=\linewidth]{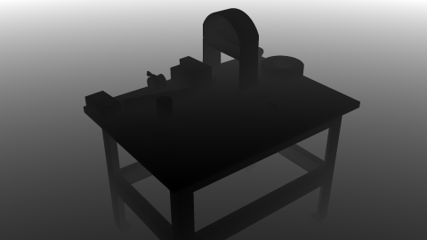}
    \caption{Depth}
  \end{subfigure}
  \begin{subfigure}{0.24\linewidth}
    \includegraphics[width=\linewidth]{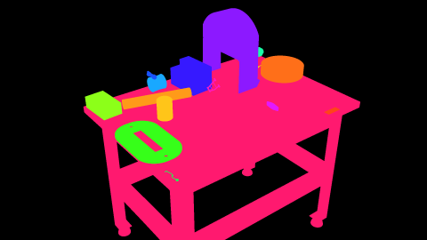}
    \caption{Masks}
  \end{subfigure}
  \begin{subfigure}{0.24\linewidth}
    \includegraphics[width=\linewidth]{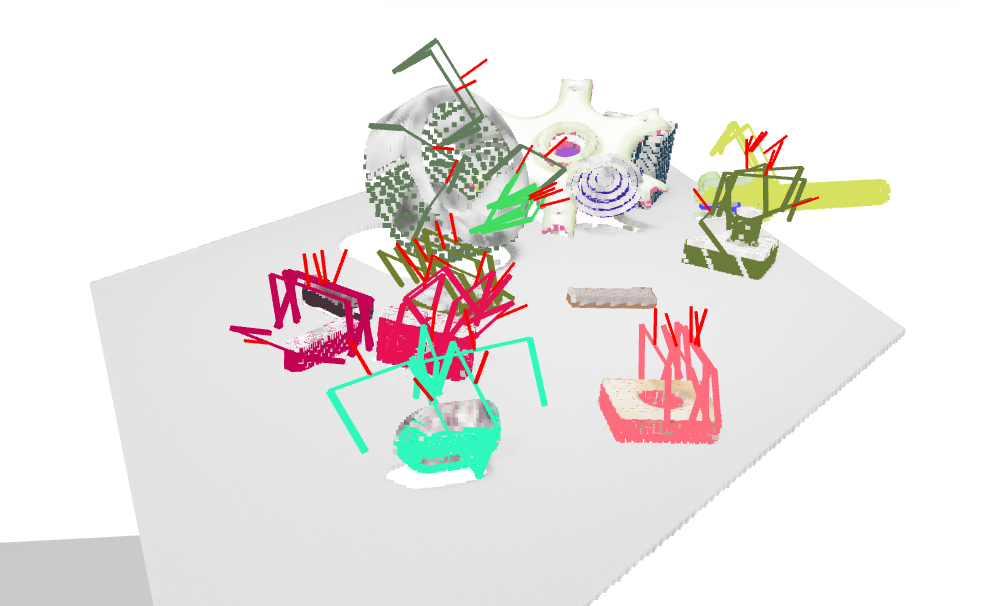}
    \caption{Grasp+Shape}
  \end{subfigure}
  \caption{\textbf{GraspIT Annotation Modalities.} RGB, metric depth  with sensor noise, per-instance segmentation, and validated 6-DoF grasp candidates overlaid on the 3D point cloud in camera coordinates.}
  \label{fig:scene_sampler_labels}
\end{figure*}

Every GraspIT frame carries the full annotation stack shown in Fig.~\ref{fig:scene_sampler_labels}: \textbf{RGB-D}: photorealistic color image and metric depth with structured-light noise applied to match real sensor characteristics~\cite{rasim}. \textbf{Instance segmentation}: per-object binary masks derived directly from the Isaac Sim scene graph. \textbf{6-DoF object pose}: ground-truth $SE(3)$ transform for every visible object. \textbf{Camera Calibration}: per view extrinsic and intrinsic calibrations. \textbf{3D Point Cloud}: full CAD surface geometry. \textbf{Physical properties}: geometric centroid (center-of-mass proxy under uniform density), spatial extent, visibility (\%), and convex-hull volume fraction.\textbf{Validated 6-DoF grasps}: slip-test quality scores per candidate grasp with translation and orientation (the pose origin is placed between the fingertips). \textbf{Grasping and Slip-Test Trajectories}: each grasp attempt has a robot trajectory with respect to the reached slip-test stage, which could be followed in Isaac Sim with full demonstration capturing (robot states, camera observations, and action commands) for policy learning. 

\subsection{Real Branch: Capture, Scan, and Register}
\label{sec:realbranch}

\begin{figure}[h]
  \centering
  \includegraphics[width=\linewidth]{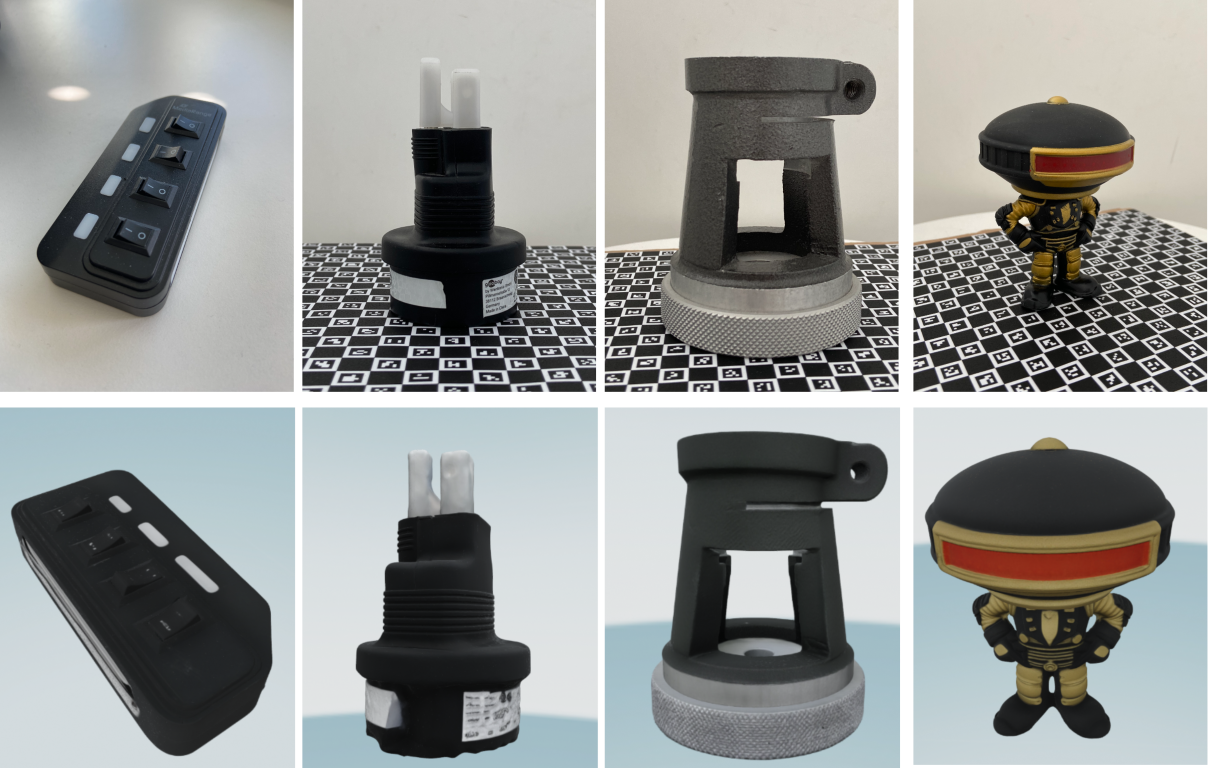}
  \caption{\textbf{Object Scan Quality.} Top row: real physical objects and scene. Bottom row: reconstructed textured meshes imported into Isaac Sim and grasp annotations.}
  \label{fig:real_branch_scan}
\end{figure}

\paragraph{Object scanning.} Physical objects are 3D scanned using TRELLIS~\cite{xiang2025native}, a 
generative image-to-3D reconstruction model. A set of high-resolution RGB images is captured under controlled 
lighting from multiple viewpoints; TRELLIS reconstructs a high-fidelity textured 3D asset with PBR material attributes (base colour, metallic, roughness). Absolute scale is recovered by fitting the reconstructed mesh to real-world scene geometry via multi-view ICP using metric camera poses from the robot's high precision forward kinematics. GraspIT initial release includes 100 real objects scanned via this pipeline (See Fig. \ref{fig:real_branch_scan} for example illustration).

\paragraph{Real Scene capture.} A UR-5 arm equipped with an Intel RealSense D405 ($1280{\times}720$) executes a hemispherical scanning trajectory of 256 predefined joint configurations. Eye-in-hand calibration provides the rigid transform $T_\text{cam}^\text{EE}$, so every camera pose in world coordinates is obtained from forward kinematics alone.

\subsection{Robot School: Physically Validated Grasp Annotation}
\label{sec:robotschool}

\begin{figure}[t]
  \centering
  \includegraphics[width=\linewidth]{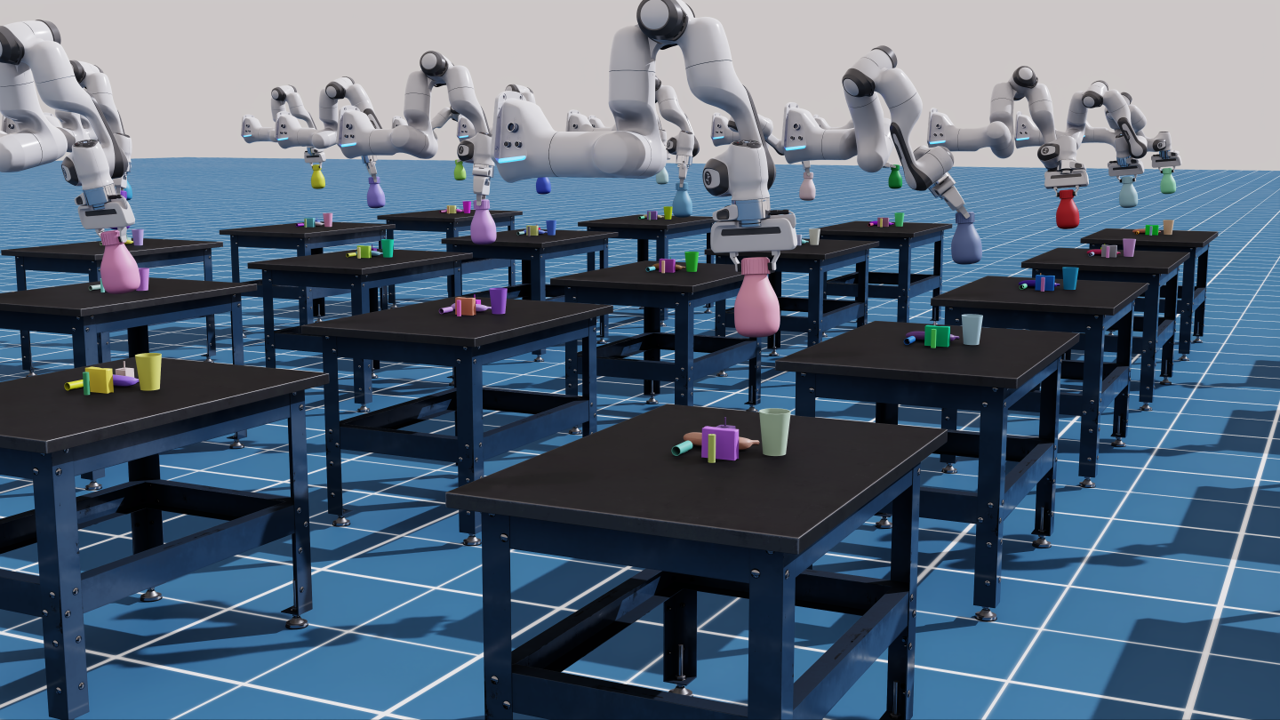}
  \caption{\textbf{Robot School.} Parallel Franka Panda instances execute grasp candidates through four sequential slip tests. The robot is spawned at a ``perfect reach'' position 90$^\circ$ tilted above the object plane to ensure reachable paths exist for all validated candidates.}
  \label{fig:robotershool}
\end{figure}

\paragraph{Candidate generation and filtering.} For every object, grasp candidates are sampled by antipodal point sampling on the down-sampled surface point cloud (1000 points). From all initial points, geometric antipodal filtering (${\sim}10\%$ pass) and collision checking with Franka Panda trajectory planning via cuRobo~\cite{sundaralingam2023curobo} and Lula~\cite{nvidia_lula} discards kinematically infeasible configurations before physics validation (passes $\sim88.54$\% of remaining) -- reducing the pool to ${\sim}307$ candidates per object. We use the Franka Hand gripper with $50$mm finger length and $80$mm opening width. The center grasping point between the fingertips marks the pose SE(3) origin, and the orientation defines the approach direction of the gripper. 

\paragraph{Four-stage slip test.} Surviving candidates enter the \emph{Robot School} (Fig.~\ref{fig:robotershool}): parallel Franka Panda instances in Isaac Sim simultaneously execute each grasp through four sequential perturbations. Position slip $>2$\,cm or orientation slip $>10°$ at any stage constitutes failure, yielding the continuous quality score defined in Eq.~\ref{eq:quality_score}. The four stages are: \textbf{1) Trajectory + Close:} trajectory to gripper fingers antipodal force-closure contact; failure indicates a geometric collision, insufficient contact, or slip. \textbf{2) Lift:} 20\,cm vertical displacement under gravity; failure indicates insufficient friction force between the fingertips, the Franka Hand closure-force ($70$N), and the object material. We use the default friction coefficient of the Isaac Sim material for all objects and the fingertips ($\mu_s 1.0$), and Isaac Sim scales object weight linearly with object volume (1000 kg/m$^3$). \textbf{3) Horizontal oscillation:} linear perturbation in the horizontal plane; exposes rotational instability. \textbf{4) Pendulum swing:} rotation of joint~5; tests resistance to moment arms created by off-center mass.

This staged protocol captures failure modes invisible to static force-closure: an object that passes vertical lift may still slip under pendulum torque, and the quality score encodes \emph{which} failure mode occurred (Fig~\ref{fig:robotershool}) and to what extent slip occurred.

\begin{figure*}[h]
  \centering
  \includegraphics[width=1.0\linewidth]{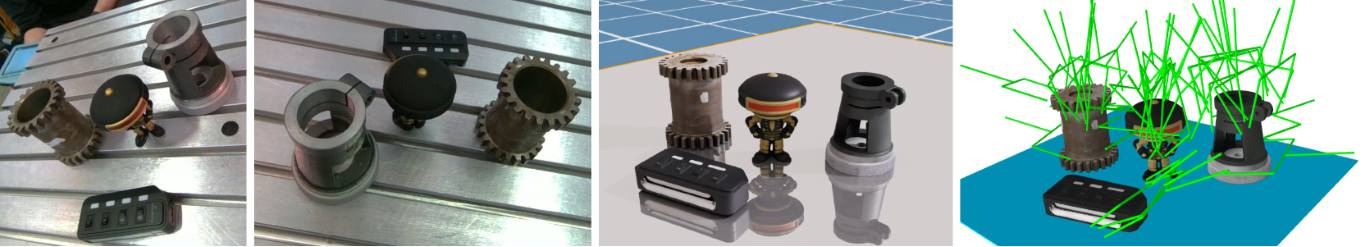}
  \caption{\textbf{Real$\leftrightarrow$Sim.} Illustration of real Scenes brought to simulation to become grasp annotated.}
  \label{fig:realscenes}
\end{figure*}

\paragraph{Annotation transfer.} In the top-down reference frame, ground-truth object poses are obtained via ICP-refined CAD fitting and propagated to all 256 real views through relative camera transforms. The registered scene is then imported into Isaac Sim; additional synthetic views are rendered at the real camera intrinsics, and per-instance segmentation masks are back-projected to real frames, and grasps are annotated in the slip test (see Fig.~\ref{fig:realscenes}).

\paragraph{Demonstration Ready.} For policy learning (behavior cloning), successful demonstrations are required, while failures can be used for reinforcement, out-of-distribution detection, and data curation (filtering and uncertainty estimation)~\cite{pi06}. The GraspIT Robot School is capable of capturing and streaming robot states, camera observations, and goal conditioning into generated successful and failure demonstrations. In combination with photorealistic Real$\leftrightarrow$Sim assets, this yields an unlimited exploration foundation that can be finetuned with minimal real data demonstrations to address real applications efficiently and at scale.

\subsection{Dataset Statistics}
\label{sec:statistics}
Tab.~\ref{tab:graspit_stats} summarizes the GraspIT initial release. The dataset comprises \textbf{1,035 simulated scenes}, each rendered at $1920{\times}1080$ from \textbf{256 hemispherical camera poses}, yielding \textbf{264,960} annotated frame sets (each with RGBD, Mask, etc). Scenes contain $13.9 \pm 2.7$ objects on average (range 8--19), from 7,396 randomly drawn unique object identities from 10k ABC object pool ($\sim73\%$ of ABC pool used).

In total, \textbf{2,275,352 grasp candidates} were evaluated by the Robot School, of which \textbf{1,887,185 (82.94\%)} are annotated \emph{good} ($s \geq 0.50$) and \textbf{388,167 (17.06\%)} \emph{bad} ($s < 0.50$). The mean score across all grasps is $\bar{s} = 0.72 \pm 0.30$; good grasps average $\bar{s}_\text{good} = 0.85$ while bad grasps average $\bar{s}_\text{bad} = 0.13$.

We use Nvidia RTX6000 GPUs to generate GraspIT. A single GPU can create a batch of 100 scenes in less than a week's compute. Generating GraspIT takes $\sim2-3$month on a single GPU (with the majority of the time being used for robot trajectory computation and grasp evaluation). 

The pipeline ablation stats (Tab.~\ref{tab:graspit_stats}) show that 88.54\% of attempted candidates obtain a planned trajectory; of those, 88.48\% survive horizontal perturbation, and 94.26\% survive the pendulum test. The largest attrition is at the trajectory-planning stage (11.46\% of all candidates receive $s{=}0.00$), confirming that robot-reachability filtering is a genuine bottleneck.

\begin{table}[t]
  \centering
  \caption{\textbf{GraspIT v1 Statistics.}
    ``Good'': $s \geq 0.50$; ``Full success'': $s = 1.00$. Penalised grasps have a base stage passed but final $s$ pushed below 0.50 by excessive slip (Eq.~\ref{eq:quality_score}).}
  \label{tab:graspit_stats}
  \resizebox{\linewidth}{!}{%
  \begin{tabular}{l|r}
    \toprule
    \textbf{Property} & \textbf{Value} \\
    \midrule
    Sim scenes                              & 1\,035 (256 sim views each) \\
    Real scenes                             & 100 (256 real + 256 synth views each) \\
    Sim annotated frame sets                & (1\,035 + 100)$\times$256 = 290\,560  \\
    Real annotated frame sets               & 100$\times$256 = 25\,600 \\
    Total annotated frames                  & 290\,560 + 25\,600  = 316\,160  \\
    Image resolution                        & 1920$\times$1080 (Sim) 1280$\times$720 (Real)  \\
    Unique objects                          & 7\,396 (Sim) + 100 (Real) \\
    Objects per scene                       & $13.9 \pm 2.7$\quad(range 8--19) \\
    \midrule
    \textbf{Grasp Pipeline}                 & \\
    Total candidates passed force-closure   & 2\,275\,352 \\
    \quad w/ trajectory                     & 2\,014\,689 (88.54\%) \\
    Total Good grasps ($s \geq 0.50$)       & 1\,887\,185 (82.94\%) \\
    Total Bad grasps ($s < 0.50$)           & 388\,167 (17.06\%) \\
    Mean Grasps per Object\& Scene          & 307$\pm$649 \& 2198$\pm$1866\\
    Good \& Bad Grasps per Scene            & 1823$\pm$1612 \& 375$\pm$423 \\ 
    Good \& Bad Grasps per object           & 255$\pm$567 \& 52$\pm$155 \\
    Mean score $\bar{s}$ (Good \& Bad)      & 0.72 $\pm$ 0.30 (0.85 \& 0.13) \\    
    \midrule
    No Trajectory to closure ($s = 0.00$)   & 260\,663 (11.46\%) \\
    Fail lift ($s = 0.25$)                  & 33\,029 ~~(1.45\%) \\
    Penalized large slip  ($s < 0.5$)       & 94\,475 ~~(4.15\%) \\
    Fail horizontal ($s = 0.50$)            & 167\,549 ~~(7.36\%) \\
    Fail pendulum ($s = 0.75$)              & 883\,733 (38.84\%) \\    
    Full success ($s = 1.00$)               & 835\,903 (36.74\%) \\
    \bottomrule
  \end{tabular}}
\end{table}

The 82.94\% good-grasp rate reflects the stringency of slip-test validation relative to analytic metrics. In particular, the penalized ``bad'' bucket (94,475 grasps) demonstrates the value of Eq.~\ref{eq:quality_score}: these candidates nominally pass two or more stages yet exhibit excessive slip, making them unreliable in practice; a pure stage-count score would incorrectly label them as good. The fine granular grasp scoring system allows for future work on novel grasp critic functions that can anticipate the quality of a physically plausible grasp based on, e.g., the generated object properties of center-of-mass, shape extend, or volume.   

Tab.~\ref{tab:grasp_comparison} compares GraspIT's annotation strategy against competing datasets.
\begin{table}[t]
  \centering
  \caption{\textbf{Grasping Dataset Comparison.} PJ: parallel jaw; Phys.~val.: physics-validated; Reach.: robot reachability filtered; Obs.: visual observations included; Sim$\leftrightarrow$Real: annotations transferable both ways.}
  \label{tab:grasp_comparison}
  \resizebox{\linewidth}{!}{%
  \begin{tabular}{l|r|r|c|c|c|c|c}
    \toprule
    \textbf{Dataset} & \textbf{\#Obj.} & \textbf{\#Grasps}
      & \textbf{Phys.} & \textbf{Reach.}
      & \textbf{Scores} & \textbf{Obs.}
      & \textbf{Sim$\leftrightarrow$Real} \\
    \midrule
    Cornell~\cite{cornellgrasp}   & 240   & 5.1k  & \ding{55} & \ding{55} & 1 & \ding{51} & \ding{55} \\
    Jacquard~\cite{jacquard}      & 11k   & 1.1M  & \ding{55} & \ding{55} & 1 & \ding{51} & \ding{55} \\
    GraspNet-1B~\cite{GraspNetDs} & 88    & 1.1B  & \ding{55} & \ding{55} & 10 & \ding{51} & \ding{55} \\
    ACRONYM~\cite{acronym}        & 8.8k  & 17.7M & $\approx$ & \ding{55} & 2  & \ding{55} & \ding{55} \\
    \textbf{GraspIT (ours)}       & 7.4k & 2.3M  & \ding{51} & \ding{51} & 5  & \ding{51} & \ding{51} \\
    \bottomrule
  \end{tabular}}
\end{table}

\section{Discussion}
\label{sec:discussion}
\paragraph{Gaps closed.} GraspIT is the first dataset to simultaneously provide: photorealistic RGBD observations at scale; robot-reachability-filtered grasps; staged, physically-validated quality scores with explicit slip-penalty calibration (Eq.~\ref{eq:quality_score}); complete 3D object shape, including occluded geometry; and a bidirectional Real$\leftrightarrow$Sim registration link (Tab.~\ref{tab:dataset_landscape},~\ref{tab:grasp_comparison}).

\paragraph{Score design and bad-grasp utility.} The 17.06\% bad-grasp fraction (388,167 candidates) is a deliberate feature rather than a limitation. The penalized bucket (94,475 grasps that pass two or more stages yet score $< 0.50$ due to excessive slip) provides particularly informative hard negatives for training grasp-quality regressors: these candidates look geometrically plausible but fail under realistic perturbation—exactly the failure mode that force-closure metrics miss.  Out-of-distribution detection, reinforcement-learning negative examples, and data curation algorithms~\cite{pi06} all benefit from this granular annotation.

\paragraph{New capabilities: Pretraining and VLA grounding.} Recent analyses~\cite{yang2026GeoVLAs, knowledgeinsulation} show that current VLAs inherit a depth-estimation deficit of ${\sim}2{\times}$ relative to geometric foundation models at the encoder level --- a gap that downstream fine-tuning can only partially close. GraspIT's physically grounded labels (shape, centroid, grasp quality) are designed to reduce this deficit when used as a pretraining stage for VLAs~\cite{pi0_paper,openvla,groot} and diffusion policies~\cite{diffusionpolicy,dp3} before task-specific fine-tuning --- enabling the grounding of the encoder stages to understand object properties and affordance. 

\paragraph{Scalability via the ABC pool.} The ABC CAD library~\cite{abcDS} (${\sim}10$k meshes) ensures highly diversity. With 7,396 unique objects across 1,035 sim scenes (mean 13.9 objects/scene) and 100 real objects and scenes, each new generation run introduces novel combinations that prevent object-identity overfitting. This scalability stands in contrast to GraspNet (88 objects) or YCB-Video (21 objects), where models risk learning object-specific shortcuts.

\paragraph{Annotation upgrade for existing datasets.} The Real$\leftrightarrow$Sim pipeline can \emph{enhance} rather than replace existing datasets. BOP sub-datasets and GraspNet scenes with known object meshes can be imported into Isaac Sim and their grasp candidates generated or validated via the Robot School, converting analytic annotations to slip-test-validated labels while preserving the original real RGBD observations.

\paragraph{Limitations.} \emph{(i) No Downstream Evaluation.} At this stage, we do not provide a downstream evaluation of GraspIT as the dataset has no single purpose that has an established evaluation. A comprehensive downstream benchmark is beyond the scope of this initial release; we leave comparative evaluation to future work. \emph{(ii) Parallel-jaw gripper only.}  GraspIT covers only parallel-jaw configurations; multi-finger and suction grasps are out of scope. \emph{(iii) Single robot model.}  Only the Franka Panda arm is used for reachability filtering; transferred labels may not be directly valid for different kinematic chains. \emph{(iv) No task demonstrations.}  GraspIT provides grasp-level annotations with trajectories, but no full task-level demonstration trajectories yet. However, the trajectories can be recaptured with the full robot state, camera observation, and action space tracking. Extending the Robot School infrastructure to pick-and-place demonstrations with language annotation is a direct next step and can be adopted by GraspIT users depending on their setup and modality needs. \emph{(v) generated physical properties}. The center-of-mass is derived from CAD geometry, while material dependent friction and volume are ignored. To further improve the current approximations, real object weighting, material simulation, and volume estimations have to be implemented into the object corpus to narrow the gap between the simulated manipulations and real world behaviors. \emph{(vi) Real$\leftrightarrow$Sim scan and registration offset} scanning, scaling, and registration of the real objects create inevitable unique and method- or hand-precision-dependent offsets that are inherent in the projected annotations. Truly measuring them in real world evaluations is impossible and can only be approximated with high precision scanning that exceeds our project budget, or their severity can be reflected in the future work downstream task evaluations and pretraining ablations. 

\paragraph{Future work.}
Extending the Robot School to full manipulation demonstrations (grasp~$\to$~lift~$\to$~place) with language annotation would create a grounded policy-learning dataset bridging GraspIT's physical labels and demonstration datasets such as DROID~\cite{droid} and Open X-Embodiment~\cite{openx}.  Additional robot models, new sensor noise profiles, and multi-camera configurations can be parameterized at the Isaac Sim level without changes to the export tooling.  We release GraspIT initially as an open foundation starting point and invite the community to contribute new scenes, objects, and annotations.

\section{Conclusion}
\label{sec:conclusion}
We presented \textbf{GraspIT}, a living dataset and generation system for physically grounded robot grasping manipulation.  By combining photorealistic Isaac Sim rendering with a four-stage physical slip test, robot-reachability filtering via cuRobo, a continuous quality score that additionally penalizes excessive positional and orientational slip (Eq.~\ref{eq:quality_score}), and a bidirectional Real$\leftrightarrow$Sim registration loop, GraspIT provides annotation coverage that no existing dataset achieves simultaneously.

The initial release comprises $\sim316$k annotated RGB-D frames across 1,035 simulated and 100 real scenes spanning $\sim7.4$k unique object geometries, with $\sim$2.3M slip-test-validated grasp candidates and fine grained scores $[0, 1]$ and slip-test stage pass annotations — of which \textbf{82.94\%} are annotated as good ($s \geq 0.50$) and the remaining \textbf{17.06\%} provide graded negative examples, including a penalized bucket of physically ambiguous near-misses (94,475 (4.15\%) candidates). The mean quality score of $\bar{s} = 0.85$ across all validated ``good'' grasps reflects the genuine physical rigor of the staged slip-test, in contrast to the binary or force-closure-only labels of prior datasets.

With our staged slip-test, we demonstrated again that antipodal force-closure is not sufficient to yield high-quality ``good'' grasp candidates. Within our slip-tests, we showed that adding path planning for physical plausibility (11.46\% filtered from force-closure accepted candidates) and staged slip-test filtering (grasped but slip filtered 17.06\% from force-closure accepted candidates) yields a better approximation for ``good'' grasp candidates than force-closure alone. Moreover, filtering ``bad'' grasping candidate targets adds the opportunity to investigate future work on critic functions and grasping quality aware grounding.

All tools, Docker containers, and the GraspIT release are publicly available at \url{https://github.com/KochPJ/GraspIt}.  We invite the community to contribute new objects, environments, robot models, and task demonstrations to grow the GraspIT database as a shared foundation for generalized robotic manipulation.

{
    \small
    \bibliographystyle{ieeenat_fullname}
    \bibliography{main}
}

\end{document}